\documentclass[twoside]{article}
\usepackage{aistats2017}
\usepackage[utf8]{inputenc} % allow utf-8 input
\usepackage{amsfonts}       % blackboard math symbols
\usepackage{nicefrac}       % compact symbols for 1/2, etc.
\usepackage{natbib}
\usepackage{todonotes}
\usepackage{amsmath,amssymb}
\usepackage{dsfont}

\begin{document}

\runningtitle{Integral Transforms from Finite Data}


\twocolumn[

\aistatstitle{Integral Transforms from Finite Data: An Application of Gaussian Process Regression to Fourier Analysis}

\aistatsauthor{ Luca Ambrogioni \And  Eric Maris }

\aistatsaddress{ Radboud University } 
]

\begin{abstract}
Computing accurate estimates of the Fourier transform of analog signals from discrete data points is important in many fields of science and engineering. The conventional approach of performing the discrete Fourier transform of the data implicitly assumes periodicity and bandlimitedness of the signal. In this paper, we use Gaussian process regression to estimate the Fourier transform (or any other integral transform) without making these assumptions. This is possible because the posterior expectation of Gaussian process regression maps a finite set of samples to a function defined on the whole real line, expressed as a linear combination of covariance functions. We estimate the covariance function from the data using an appropriately designed gradient ascent method that constrains the solution to a linear combination of tractable kernel functions. This procedure results in a posterior expectation of the analog signal whose Fourier transform can be obtained analytically by exploiting linearity. Our simulations show that the new method leads to sharper and more precise estimation of the spectral density both in noise-free and noise-corrupted signals. We further validate the method in two real-world applications: the analysis of the yearly fluctuation in atmospheric CO\textsubscript{2} level and the analysis of the spectral content of brain signals.
\end{abstract}

\section{Introduction}
The Fourier transform is perhaps the most important mathematical tool for the analysis of analog signals. In order to be processed with digital computers, analog signals need to be sampled at a finite number of time points. From the samples, the Fourier transform of the signal is usually estimated using the discrete Fourier transform (DFT). However, the Nyquist–Shannon sampling theorem states that the DFT provides an unbiased estimate if and only if the underlying signal is periodic and contains no power above the Nyquist frequency \citep{rabiner1975theory}. Unfortunately, estimating the Fourier transform of signals that do not respect these properties is an intrinsically ill-posed problem as there are infinitely many functions that perfectly fit any finite set of samples. From a Bayesian perspective, ill-posed problems can be solved by assigning a prior probability distribution to the underlying functional space \citep{idier2013bayesian}. Gaussian process (GP) priors have gained substantial popularity in these kinds of applications due to their flexibility, robustness and analytical tractability \citep{rasmussen2006gaussian}. With this paper we introduce the use of GP regression for estimating the Fourier transform (or any other linear transform) of analog signals from a finite set of samples. The estimation procedure assumes neither periodicity nor discreteness of the signal and outputs a closed-form function expressed as a linear combination of tractable kernels. This latter feature is particularly important since it allows to analytically perform a wide range of further analyses using closed-form expressions. This reduces the impact of numerical errors and instabilities on the analysis pipeline. We estimate the GP covariance function using a constrained gradient ascent method that maintains the analytical tractability while being able to analyze arbitrarily complex signals with high extrapolation and interpolation performance.

\subsection{Related works}
Bayesian methods have become very influential in the field of spectral estimation \citep{gregory2001bayesian}. Several nonparametric Bayesian  approaches have been applied to the estimation of the spectral density of stochastic signals \citep{carter1997semiparametric, gangopadhyay1999estimation, liseo2001bayesian}. These methods are based on an asymptotic approximation of the distribution of the periodogram of discrete-time signals \citep{whittle1957curve}. Recently, some work has been done on the use of GP regression for stochastic spectral estimation of discretely sampled analog signals. The GP spectral mixture (GP-SM) approach models the spectral density by fitting the parameters of a small number of Gaussian functions \citep{wilson2013gaussian}. A more flexible alternative is the Gaussian Process Convolution Model (GPCM), which is based on a two-stage generative model where the signal is assumed to be generated by convolving a white noise process with a filter function sampled from a GP \citep{wilson2013gaussian}. The estimates obtained using GPCM and GP-SM do not directly provide an estimate of the Fourier transform since the spectral density does not contain phase information. In our method, we learn the spectral density using a new gradient ascent method that shares the nonparametric flexibility of GPCM while being significantly simpler. This simplicity comes at the price of neglecting the uncertainty about the spectral density estimate. The most important feature of this learning approach is that the resulting point-estimate is expressed as a linear combination of tractable kernel functions. This is a requirement for our Fourier estimation procedure. This Fourier estimation procedure can be seen as a generalization of the GP quadrature method, which uses GP regression for numerical integration of definite integrals \citep{o1991bayes}. In this approach, the integrand is assumed to be sampled from a GP distribution and evaluated on a finite set of points. Importantly, under these assumptions, the posterior distribution of the integral can be obtained in closed-form. 

\section{Background}
The Fourier transform of a (real or complex valued) function $f(t)$ is defined as follows:
\begin{equation}
\mathfrak{F}\big[f(t)\big](\xi) = f(\xi) = \frac{1}{{2 \pi}} \int_{-\infty}^{+\infty} e^{-i \xi t} f(t) dt ~,
\label{eq: Fourier transform}
\end{equation}
where $e^{-i \xi t} = \cos{\xi t} - i \sin{\xi t}$ is a complex valued sinusoid. The Fourier transform is a linear operator, meaning that $\mathfrak{F}\big[a f(t) + b g(t)\big](\xi) = a \mathfrak{F}\big[f(t)](\xi) + b \mathfrak{F}\big[g(t)](\xi)$. We can interpret the Fourier transform as a special case of linear integral transform. A general linear integral transform has the following form:
\begin{equation}
\mathfrak{I}_{\mathcal{A}}\big[f(t)\big](s) = \int_{b}^{a} \mathcal{A}(s,t) f(t) dt ~,
\label{eq: Integral transform}
\end{equation}
where the bivariate function $\mathcal{A}(s,t)$ is the kernel of the transform. The limits of integration, $a$ and $b$, can be finite or infinite.

\subsection{Gaussian Process Regression}
GP methods are popular Bayesian nonparametric techniques for regression and classification. A general regression problem can be stated as follows:
\begin{equation}
y_t = f(t) + \epsilon_t~,
\label{eq: regression problem}
\end{equation}
where the data point $y_t$ is generated by the latent function $f(t)$ plus a zero-mean noise term $\epsilon_t$ that we will assume to be Gaussian. The main idea of GP regression is to use an infinite-dimensional Gaussian prior (a GP) over the space of functions $f(t)$. This infinite-dimensional prior is fully specified by a mean function, usually assumed to be identically equal to zero, and a covariance function $\mathfrak{K}(t,t')$ that determines the prior covariance between two different time points. The posterior distribution over $f(t)$ can be obtained by applying Bayes theorem. Given a set of training points $(t_k,y_k)$, it can be proven that the posterior expectation $m_{f}(t)$ is a finite linear combination of covariance functions:
\begin{equation}
m_{f}(t) = \sum_{k} w_k \mathfrak{K}(t,t_k)~,
\label{eq: posterior expectation}
\end{equation}
where the weights are linear combinations of data points
\begin{equation}
w_k = \sum_{j} A_{kj} y_j~.
\label{eq: weights}
\end{equation}
In this expression the matrix $A$ is given by the following matrix formula:
\begin{equation}
A = (K + \lambda I)^{-1}~,
\label{eq: filter matrix}
\end{equation}
where $\lambda$ is the variance of the sampling noise and the matrix $K$ is obtained by evaluating the covariance function for each couple of time points:
\begin{equation}
K_{jk} = \mathfrak{K}(t_j,t_k)~.
\label{eq: covariance matrix}
\end{equation}
The derivation of these results is given in \citep{rasmussen2006gaussian}. 

\section{Computing Integral Transforms Using GP Regression}
One of the most appealing features of GP regression is that, while the training data are finite and discretely sampled, the posterior expectation is defined over the whole time axis. Furthermore, Eq.~\ref{eq: posterior expectation} shows that this expectation is a linear combination of covariance functions. From this linearity, it follows that every integral transform of $m_{f}(t)$ can be calculated as a linear combination of the integral transform of the covariance functions $\mathfrak{K}(t,t_k)$: 
\begin{align}
\mathfrak{I}_{\mathcal{A}}\big[m_{f}\big](s) &= \int_{b}^{a} \mathcal{A}(s,t) \bigg[ \sum_{k} w_k \mathfrak{K}(t,t_k) \bigg] dt \label{eq: GP integral transform} \\ \nonumber
&= \sum_{k} w_k \int_{b}^{a} \mathcal{A}(s,t) \mathfrak{K}(t,t_k) dt ~.
\end{align}
Clearly, this transform is well-defined as far as the transform $\int_{b}^{a} \mathcal{A}(s,t) \mathfrak{K}(t,t_k) dt$ exists. The special case where the integral operator is a simple definite integral has been applied to numerical integral analysis and is known as the GP quadrature rule \citep{o1991bayes}:
\begin{equation}
\int_{b}^{a} f(t) dt \approx \sum_{k} w_k \int_{b}^{a} \mathfrak{K}(t,t_k) dt ~.
\label{eq: GP quadrature}
\end{equation}

In the case of the Fourier transform, Eq.~\ref{eq: GP integral transform} becomes:
\begin{align}
\mathfrak{I}_{F}\big[m_{f}\big](\xi) = \sum_{k} w_k\frac{1}{{2 \pi}} \int_{-\infty}^{+\infty} e^{-i \xi t} \mathfrak{K}(t,t_k) dt~.
\label{eq: GP fourier transform}
\end{align}
This expression further simplifies when $\mathfrak{K}(t,t_k)$ is stationary, meaning that $\mathfrak{K}(t,t') = \mathfrak{K}(t - t',0)$. In this case, we can use the Fourier shift theorem and obtain:
\begin{align}
&\Bigg(\sum_{k} w_k e^{-i \xi t_k} \Bigg) \Bigg( \frac{1}{2 \pi} \int_{-\infty}^{+\infty} e^{-i \xi t} \mathfrak{K}(t,0) dt~ \Bigg) \\ 
\label{eq: stationary GP fourier transform} \nonumber
&= \mathfrak{K}(\xi,0) \sum_{k} w_k e^{-i \xi t_k}~.
\end{align}
where the rightmost factor in the right hand side is proportional to the DFT of the GP weights, as defined in Eq.~\ref{eq: weights}. This result hints to a deep connection between the GP Fourier approach and the classical DFT approach based on the Nyquist–Shannon sampling theorem. This connection will be made explicit in section \ref{sec: theoretical considerations}. 

\subsection{Hierarchical Covariance Learning}
The aim of this subsection is to introduce a hierarchical Bayesian model that allows to estimate the GP covariance function from the data using a MAP estimator. We restrict our attention to stationary covariance functions, i.e. covariance functions that solely depend on the difference between the time points $\tau = t' - t$. We construct the hierarchical model by defining a hyper-prior for the the spectral density $\mathfrak{S}(\xi)$, defined as the Fourier transform of the covariance function:
\begin{equation}
\mathfrak{S}(\xi) = \frac{1}{2 \pi}\int_{-\infty}^{+\infty} \mathfrak{K}(\tau) e^{-i \xi \tau} d\tau ~.
\label{eq: spectral density}
\end{equation}
Since the Fourier transform is invertible, an estimate of the spectral density can be directly converted into an estimate of the covariance function. Using a GP hyper-prior on the spectral density $\mathfrak{S}(\xi)$ would be a convenient modeling choice as it easily allows to specify its prior smoothness, thereby regularizing the estimation. For example, we could use a GP hyper-prior with squared exponential (SE) kernel (covariance) function: 
\begin{equation}
\mathfrak{K}_{SE}(\xi, \xi') = e^{-\frac{(\xi - \xi')^2}{2 \sigma^2}} ~,
\label{eq:squared exponential}
\end{equation}
where the scale parameter $\sigma$ regulates the prior smoothness. Unfortunately, this is not a valid prior for the spectral density of a GP since it assigns non-zero probability to negative valued spectra which do not correspond to any valid stationary stochastic process. However, we can obtain a proper prior distribution by restricting this GP probability measure to the following positive-valued functional space:
\begin{equation}
\{ \mathfrak{s}(\xi) = \sum_{j} e^{a_j} \mathfrak{K}_{SE}(\xi, \xi_j) | a_j \in \mathds{R} \}~,
\label{eq: constrained sub-space}
\end{equation}
where $\xi_j$ are the discrete Fourier frequencies of the sampled data points. In order to obtain the likelihood of the model, we assume that each DFT coefficient only depends on the value of the spectrum corresponding to its frequency. For each frequency, the resulting likelihood functions are complex normal distributions:
\begin{equation}
\log{p\big(y_{\xi_i}|\mathfrak{s}(\xi)\big)} = -\frac{|y_{\xi_j}|^2}{\mathfrak{s}(\xi) + \lambda} -\log{\big( \pi(\mathfrak{s}(\xi) + \lambda)\big)}
\label{eq: marginal likelihood}
\end{equation}
where $y_{\xi_j}$ is the $j$-th DFT coefficient of the sampled data and $\lambda$ is the variance of the sampling noise. The assumption of conditional independence is justified by the fact that the Fourier coefficients of stationary GPs are independent random variables. Nevertheless, the assumption is not exact since the finite length of the sampling period induces correlations between the DFT coefficients. We mitigated the bias induced by this approximation by computing the DFT coefficients using a Hann taper, which reduces the correlations between DFT coefficients corresponding to distant frequencies. We calculate the maximum-a-posteriori (MAP) estimate by means of gradient ascent applied to the posterior distribution of the spectral density. The algorithm maximizes the posterior distribution with respect to the log-weights $a_j$ and therefore only finds solutions in the restricted subspace of Eq. \ref{eq: constrained sub-space}. In the log-weight space, the gradient of the (approximate) log marginal likelihood $\mathfrak{l}$ is 
\begin{equation}
\frac{\partial \mathfrak{l}}{\partial a_k} = e^{a_k} \sum_j \frac{\big( |y_{\xi_j}|^2 - (\mathfrak{s}(\xi_j) + \lambda) \big)}{ \big(\mathfrak{s}(\xi_j) + \lambda\big)^2} \mathfrak{K}_{SE}(\xi_k, \xi_j)~
\label{eq: marginal likelihood gradient}
\end{equation}
and the gradient of the (log-)prior $\mathfrak{p}$ is
\begin{equation}
\frac{\partial \mathfrak{p}}{\partial a_k} = - e^{a_k} \sum_j \mathfrak{K}_{SE}(\xi_k, \xi_j) e^{a_j} ~.
\label{eq: prior gradient}
\end{equation}
The resulting MAP estimate has the following form: 
\begin{equation}
\hat{\mathfrak{S}}(\xi) = \sum_{j} e^{h_j} \mathfrak{K}_{SE}(\xi, \xi_j)~,
\label{eq:spectrum estimate}
\end{equation}
where $h_j$ are the optimized log-weights. Finally, our point estimate of the covariance function is obtained by applying the inverse Fourier transform to the MAP estimate of the spectral density:
\begin{align}
\label{eq: covariance estimate}
\hat{\mathfrak{K}}(\tau) &= \int_{-\infty}^{+\infty} e^{i \xi \tau}\hat{\mathfrak{S}}(\xi) d\xi \\ \nonumber
&= \sum_{j} e^{h_j} \int_{-\infty}^{+\infty} e^{i \xi \tau}\mathfrak{K}_{SE}(\xi, \xi_j) d\xi~ \\ \nonumber
&= \sigma \sum_{j} e^{h_j} e^{-\frac{\sigma^2 \tau^2}{2} - i \xi_j \tau}.
\end{align}
This covariance function has the advantage of capturing the spectral features of the data while keeping a tractable analytic expression as a linear combination of the inverse Fourier transforms of SE kernel functions. Note that, if we have access to multiple realizations of a stochastic time series, we can learn the spectral density from the whole set of realizations simply by summing the log marginal likelihood of each realization. We will use this procedure in our analysis of neural oscillations.

\subsection{Bayes-Gauss-Fourier transform} \label{subsec: BFG transform}
We can now plug in the data-driven covariance function in our expression for the integral transform and exploit the linear structure of the covariance function by interchanging summation and integration:
\begin{equation}
\sigma \sum_{k,j} w_k e^{h_j} \int_{b}^{a} \mathcal{A}(s,t) e^{-\frac{\sigma^2 (t - t_k)^2}{2} - i \xi_j (t - t_k)} dt ~.
\label{eq: GP integral transform adaptive covariance}
\end{equation}
In the case of the Fourier transform this formula specializes to
\begin{equation}
\sum_{k,j} w_k e^{h_j} e^{-\frac{(\omega - \xi_j)^2}{2 \sigma^2} - i \omega t_k}~
\label{eq: GP Fourier transform adaptive covariance}
\end{equation}
because
\begin{equation*}
\mathfrak{F}\big[e^{-\frac{\sigma^2 (t - t_k)^2}{2} - i \xi_j (t - t_k)}\big](\omega) = \sigma^{-1} e^{-\frac{(\omega - \xi_j)^2}{2 \sigma^2} - i \omega t_k} ~.
\end{equation*}
We refer to the resulting transformation of the data as the Bayes-Gauss-Fourier (BGF) transform.

\section{Theoretical considerations} \label{sec: theoretical considerations}
In this section, we will lay a more rigorous mathematical foundation of the newly introduced methods. The aim is to introduce a larger theoretical framework that allows to directly compare the DFT-based methods with our new GP Fourier transform. In particular, we will show that the DTF method can be seen as a special case GP Fourier transform. 

\subsection{Fourier transform of band-limited periodic signals}
We begin by reviewing some well known results about the Fourier analysis of periodic band-limited signals. This will pave the way to a Bayesian reformulation of Fourier analysis, which we will introduce in the next subsection. Consider a vector of samples $\boldsymbol{y} = (y_{t_0},..,y_{t_N})$ obtained by evaluating an analog signal at the tuple of time points $T = (t_0,...,t_N)$. The Fourier analysis of a discretely sampled analog signal can be decomposed into two basic operations. 

First, the set of samples have to be mapped into a well-defined analog signal. This operation can be formalized by a linear operator $\mathfrak{M}:Y \rightarrow \mathcal{H}$, mapping the sample space $Y$ to an appropriate functional space $\mathcal{H}$. The operator is defined by the property 
\begin{equation}
\mathfrak{M}[\boldsymbol{y}](t_j) = f(t_j) = y_j~.
\label{eq: fitting requirement}
\end{equation}
This simply means that the values of the resulting function at the sample time points have to be equal to the samples. 

Second, the analog signal has to be mapped into its Fourier transform. From an abstract point of view, the Fourier transform can be seen as a linear operator between functional spaces $\mathfrak{F}:\mathcal{H} \rightarrow \mathcal{I}$, where the exact nature of the spaces $\mathcal{H}$ and $\mathcal{I}$ depends to the specific application. Under some regularity conditions over the functional space $\mathcal{H}$ \citep{schechter1971principles}, this second operation is unproblematic. 

Unfortunately, the first operation is intrinsically ill-posed since there is not a unique way to map a finite set of samples into an analog signal. The classical solution to this problem relies on the Nyquist–Shannon sampling theorem, which states that an analog signal can be perfectly reconstructed from the finite series of equally spaced samples $\boldsymbol{y} = (y_{-T/2},y_{-T/2 + \delta t}, ... y_{T/2 - \delta t}, ,y_{T/2})$ if and only if: 1) the signal is periodic with period $T$ and 2) it does not contain harmonic components with frequency higher than $1/(2 \delta t)$. We will denote this functional space of periodic and band-limited functions as $\mathcal{H}_{bl}$. Under these conditions, the operator $\mathfrak{M}: Y \rightarrow \mathcal{H}_{bl}$ is uniquely defined. The functional space $\mathcal{H}_{bl}$ is the span of a finite number of complex exponential basis functions
\begin{equation}
\Phi_{k}(t) = \frac{1}{N} e^{i \omega_k t}~.
\label{eq: harmonic basis functions}
\end{equation}
where $\omega_k = 2 \pi k / T$ and the integer $k$ ranges from $- N/2$ to $N/2$. Therefore, any periodic and band-limited function $f_{pb}(t) \in \mathcal{H}_{bl}$ can be expressed as follows
\begin{equation}
f_{bl}(t) = \sum_k \alpha_k \Phi_{k}(t)~.
\label{eq: PB functions}
\end{equation}
Combining Eq.~\ref{eq: PB functions} with Eq.~\ref{eq: fitting requirement}, we obtain a linear system of equations with a unique solution since the set of basis functions is linearly independent. The solution coefficients are the DFT coefficients of the samples: 
\begin{equation}
f_{bl}(t) = \sum_k \big( \boldsymbol{\phi}^{*}_k\boldsymbol{y} \big)~ \Phi_{k}(t)~,
\label{eq: DTF solution}
\end{equation}
where the vectors $\boldsymbol{\phi}_k$ are obtained by evaluating the basis functions $\Phi_{k}(t)$ at the sample time points. We can now take the Fourier transform of the function in Eq.~\ref{eq: DTF solution}, the result is:
\begin{align}
\label{eq: BL fourier transform}
\mathfrak{F} [f_{bl}(t)] &= \sum_k \big( \boldsymbol{\phi}^{*}_k\boldsymbol{y} \big)~ \mathfrak{F} [\Phi_{k}(t)] \\
 &= \sum_k \frac{\big( \boldsymbol{\phi}^{*}_k\boldsymbol{y} \big)}{N}~\delta(\xi - \omega_k)~. \nonumber
\end{align}
In the last expression, the symbol $\delta(\xi - \omega_k)$ denotes the Dirac delta function. The precise meaning of this symbol relies on the theory of distributions. Intuitively, Eq.~\ref{eq: BL fourier transform} says that all the energy of the signal is concentrated in a finite number of DTF frequencies.

\subsection{Fourier transform from a functional Bayesian viewpoint}
We will now show that the classical method we reviewed in the previous subsection is a special case of GP Fourier analysis. First of all, the problem can be integrated into a Bayesian probabilistic framework by introducing observation noise into Eq.~\ref{eq: fitting requirement} and by assigning a prior distribution over the functional space $\mathcal{H}_{bl}$. Using Gaussian observation noise and a spherical Gaussian prior distribution over the coefficients $\alpha_k$, we obtain the following Bayesian problem:
\begin{align}
\label{eq: bayesian formulation} 
& y_{t_j} \sim \mathcal{N}\big(f_{bl}(t_j)~,\lambda \big) \\
&\alpha_k \sim \mathcal{N}\big(0, \beta \big), \nonumber \\
& f_{pb}(t_j) = \sum_k \alpha_k \Phi_{k}(t_j)~, \nonumber
\end{align}
where $\lambda$ is the variance of the observation noise and $\beta$ is the variance of the prior over the coefficients. This is a Bayesian linear regression. Importantly, regardless to the value of the prior variance, the posterior distribution of the coefficients concentrate all its mass on the solution given in Eq.~\ref{eq: DTF solution} when the observation noise tends to zero. Therefore, the Bayesian problem in Eq.~\ref{eq: bayesian formulation} is a probabilistic generalization of the deterministic problem given by Eq.~\ref{eq: PB functions} and Eq.~\ref{eq: fitting requirement}.

The Bayesian linear regression in Eq.~\ref{eq: bayesian formulation} can now be reformulated as a GP regression. This generalizes our analysis to functional spaces that cannot be obtained from a finite set of basis functions, thereby allowing for more flexible and realistic prior distributions. In particular, we will work on the space of complex-valued functions whose domain is $\mathds{R}$, which we will denote $\mathcal{H}_{\Omega}$. We can now define a Gaussian probability measure over $\mathcal{H}_{\Omega} \supseteq \mathcal{H}_{bl}$ that is equivalent to the coefficient space prior distribution given in Eq.~\ref{eq: bayesian formulation}. This is achieved by constructing the following covariance function \citep{rasmussen2006gaussian}:
\begin{equation}
\mathfrak{K}_{BL}(t,t') = \beta \sum_k \Phi_{k}(t) \Phi_{k}^{*}(t')~.
\label{eq: equivalent covariance function} 
\end{equation}
Using Eq.~\ref{eq: equivalent covariance function}, we can reformulate Eq.~\ref{eq: bayesian formulation} as follows:
\begin{align}
\label{eq: GP formulation}
& y_{t_j} \sim \mathcal{N}\big(f_{bl}(t_j)~,\lambda \big)  \\
& f_{pb}(t_j) \sim \mathcal{CGP}\big(0, \mathfrak{K}_{bl}(t,t') \big)~, \nonumber
\end{align}
where $\mathcal{CGP}\big(m(t), k(t,t')$ denotes a complex circularly-symmetric GP with mean function $m(t)$ and (Hermitian) covariance function $k(t,t')$ \citep{boloix2014gaussian, boloix2015complex, ambrogioni2016complex}. The main feature of this function is that the resulting correlation between any pair of sample time points is zero (except for the first and the last time point, where the correlation is exactly equal to one). However, since the covariance function is different from zero almost everywhere, the Bayesian reformulation shows that we are still making strong assumptions about the behavior of the function outside of the set of sample time points. Strikingly, the covariance function is periodic with period $T$ and, consequently, all the functions obtained from this GP are periodic with period $T$. 

From a Bayesian point of view, determining the prior distribution based on the sample time points is rather counterintuitive since our prior knowledge of the signal should not depend on the sampling frequency and the total sampling time. Furthermore, the covariance function given in Eq.~\ref{eq: equivalent covariance function} is degenerate, meaning that it assigns non-zero probability measure only to a finite dimensional functional sub-space. This leads to some rather paradoxical situations. For example, consider two scientists that are analyzing the same radio signal, the first sampling it at $300$ kHz for a period of $1$s and the second at $300$ kHz for $1.1$s. If they use Eq.~\ref{eq: equivalent covariance function}, the resulting prior distributions will have a disjoint support, meaning that the spaces of signals that they consider possible are completely non-overlapping! 

Using Eq.~\ref{eq: GP formulation}, we can generalize the analysis to other prior distributions simply by choosing another covariance function. A possible compromise solution is to multiply the band-limited covariance function given in Eq.~\ref{eq: equivalent covariance function} by a radial basis function such as the squared exponential: 
\begin{equation}
\mathfrak{K}_{rBL}(t,t') = \beta \mathfrak{K}_{SE}(t,t'; \nu) \sum_k \Phi_{k}(t) \Phi_{k}^{*}(t')~.
\label{eq: relaxed band-limited covariance function} 
\end{equation}
Where $\nu$ denotes the length scale. This \emph{relaxed band-limited} (rBL) covariance function assigns zero correlation between any couple of sample time points but it does not enforce periodicity. The resulting GP Fourier transform of the data is obtained by plugging Eq.~\ref{eq: equivalent covariance function} into Eq.~\ref{eq: posterior expectation} and taking the Fourier transform of the resulting linear combination of translated covariance functions:
\begin{align}
\label{eq: relaxed band-limited transform} 
&\mathfrak{F}[m_f](\xi) = \beta \sum_j w_j~\mathfrak{F}[\mathfrak{K}_{rBL}(t,t_j; \nu)](\xi) \\
&= \beta \nu \Bigg(\sum_{k} e^{-\nu^2 \frac{(\xi - \omega_k)^2}{2}} \Bigg)\Bigg(\frac{1}{T} \sum_{j} w_j~e^{-i \xi t_j}\Bigg)~. \nonumber
\end{align}
Crucially, while the resulting prior is still dependent on the sample time points, the rBL GP Fourier transform does not concentrate all the energy of the signal on a, rather arbitrary, finite set of frequencies. 

Both the BL and the rBL covariance functions are designed to be maximally uninformative on the sample time points. This leads to Fourier analysis that are not biased toward a particular set of frequencies. Unfortunately, this also leads to a GP analysis that does not meaningfully extrapolate the signal beyond the sample time points. In order to be able to extrapolate without biasing a predefined set of frequencies, the covariance function has to be learned from the data. In fact, this allows to detect the periodic components of the signal and to use this information in order to extrapolate beyond the sampling range. In subsection \ref{subsec: BFG transform}, we outlined a Bayesian learning scheme based on gradient ascent. The resulting BFG covariance function given in Eq.~\ref{eq: covariance estimate} can be rewritten as follows:
\begin{equation}
\hat{\mathfrak{K}}(t,t') = N^2 \mathfrak{K}_{SE}(t,t'; 1/\sigma) \sum_k e^{h_k(\boldsymbol{y})} \Phi_{k}(t) \Phi^{*}_{k}(t')~,
\label{eq: BFG covariance function} 
\end{equation}
This reformulation shows that the BFG covariance function is a spectrally  weighted version of the rBL covariance function. 

\section{Experiments}
In this section we validate our new method on simulated and real data. We focus the validation studies on the problem of estimating the power spectrum of deterministic and stochastic signals, as this is perhaps the most common application of the Fourier transform. We compare the performance of the BFG transform with more conventional DFT-based estimators. 

\subsection{Analysis of noise-free signals}
\begin{figure}[!t]
	\centering
    	\includegraphics[width=0.45\textwidth] {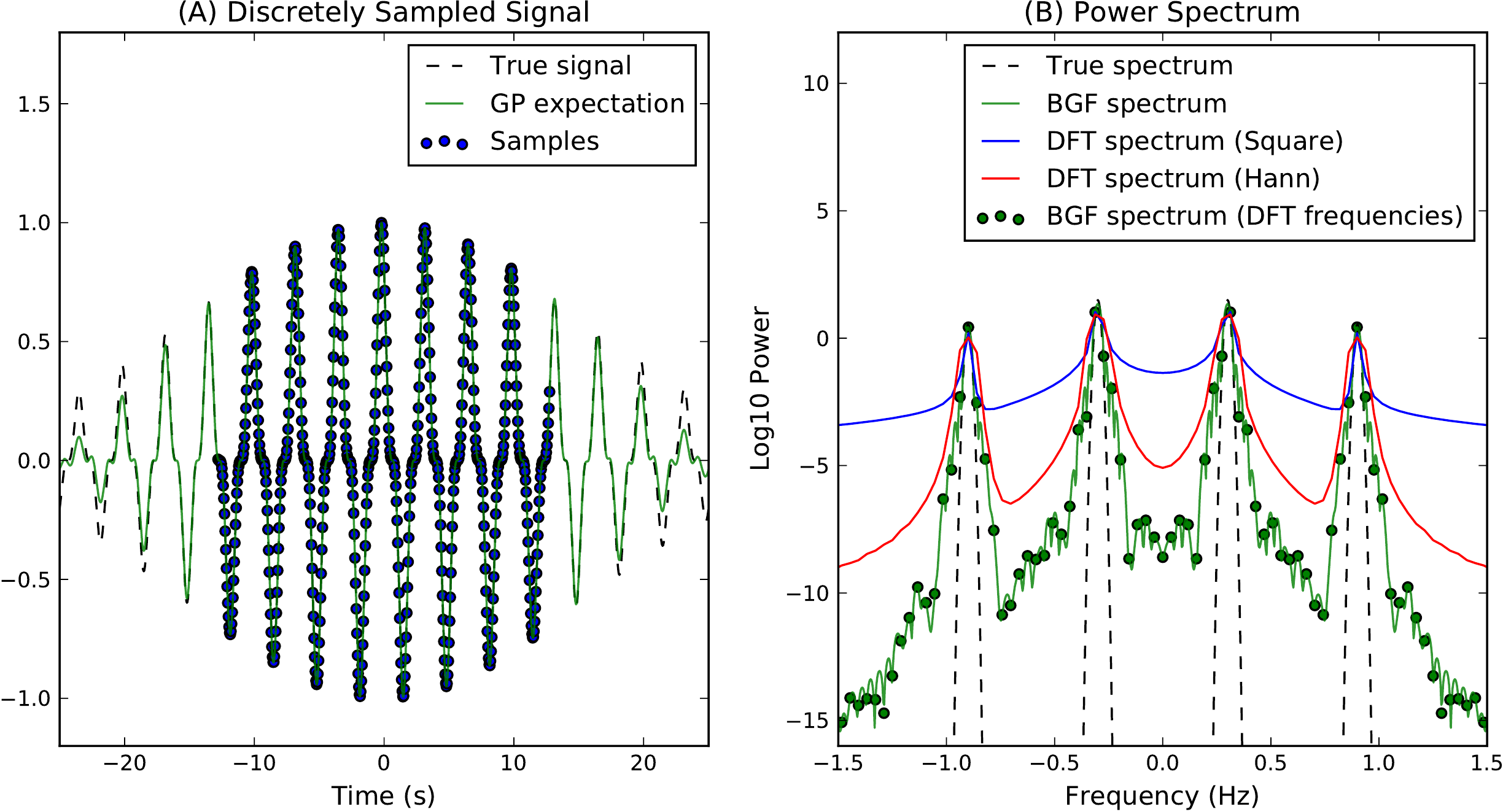}% picture filename
	\caption{Spectral estimation of a synthetic signal. A) Ground truth signal (dashed black line), sample points (blue dots) and the expected value of the GP regression (green line). B) (Log10) Power spectrum of the ground truth signal (dashed line) and spectral estimates obtained from the samples using BGF transform (green line), DTF (blue line) and DTF with Hann taper (red line).}
	\label{figure 1}
\end{figure}

We investigate the performance of our method in recovering the Fourier transform of a discretely sampled deterministic signal. As first example signal, we use the following an-harmonic windowed oscillation $ g(t) = e^{-{t^2}/{2a^2}} {\cos}^3{\omega_0 t}.$ We sampled the signal from $t_{min} = -25$ to $t_{max} = 25$ in steps of $0.01$ and with $a = 15$ and $\omega_0 = \frac{3}{5} \pi$. These samples were analyzed using the BGF transform as described in the Methods. Fig.~\ref{figure 1}A shows the result of the GP regression in the time domain. Clearly, the expected value of the GP regression (blue line) is able to extrapolate the waveform of the signal far beyond the data points. Next we compared our GP-based estimate with two more conventional estimates of the spectrum $|g(\omega)|^2$: the Discrete Fourier Transform (DFT) of the data using a square and a Hann taper. The spectra obtained using the DFT methods were normalized to have the same energy of the ground truth signal in the set of DTF frequencies. This normalization is required since the DTF-based methods assign all the energy of the analog signal to the DTF frequencies instead of spreading it into a continuous spectrum. Fig.~\ref{figure 1}B shows these spectral estimates, together with the ground truth spectrum, on a log scale. The BGF transform (green line) captures the shape and width of the four main lobes almost perfectly, despite the fact that their peaks are not fully aligned with the discrete Fourier frequencies of the sampled data (which are determined by the  signal's length). Furthermore, the BGF transform has significantly higher sidelobe suppression than the DFT estimates, up to $10^6$ higher than the DFT with Hann taper. 

\begin{figure}[!t]
	\centering
    	\includegraphics[width=0.45\textwidth] {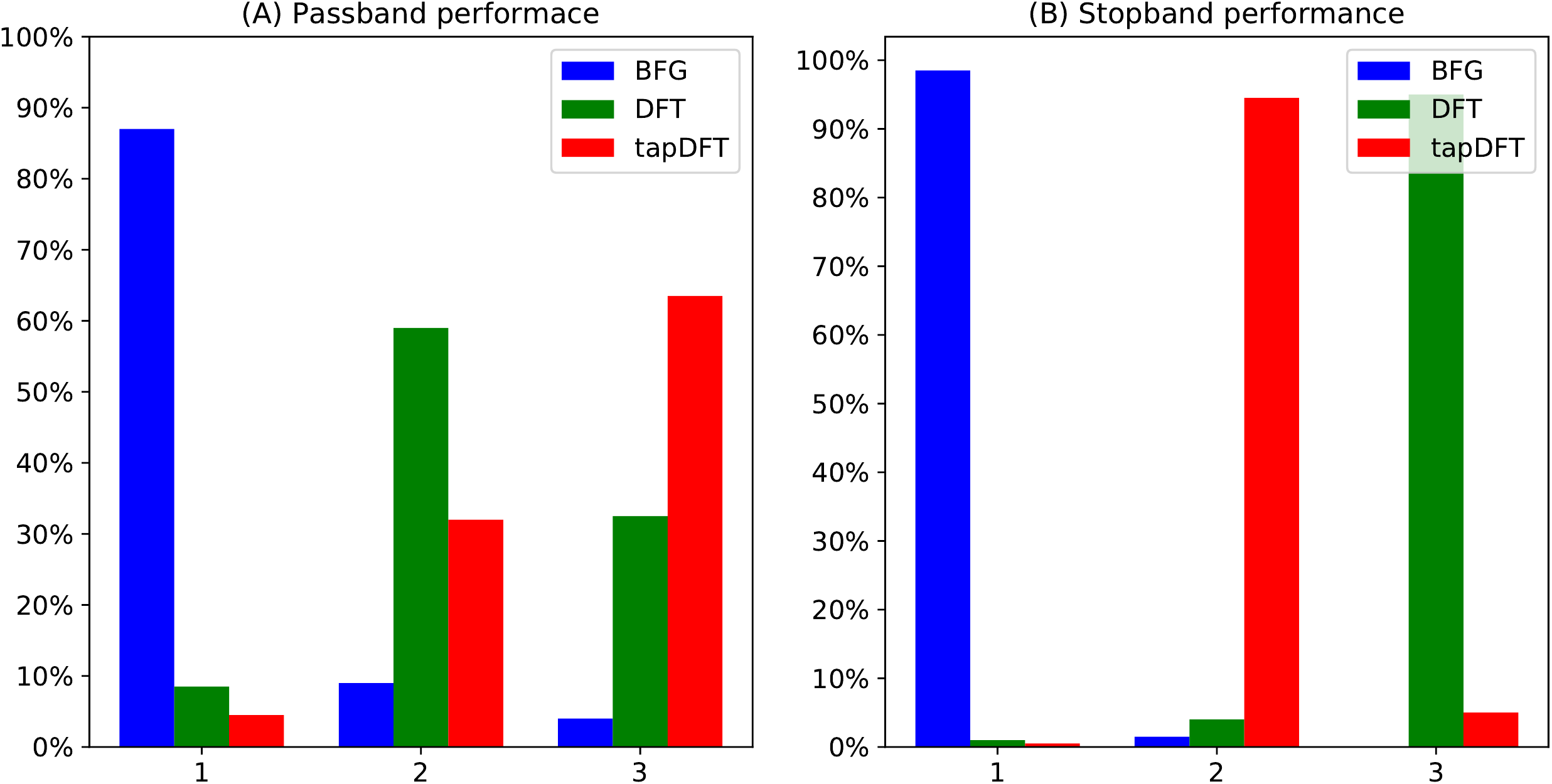}% picture filename
	\caption{Quantitative comparison (noise-free case). Ranked absolute deviations in the A) passband range ($\log_{10}g(\xi) > -6$) and B) stopband range ($\log_{10}g(\xi) < -6$)}
	\label{figure 1b}
\end{figure}

We quantitatively evaluated these observations in a simulation study. We randomly generated $200$ ground truth signals of the form $ g(t) = e^{-{t^2}/{2a^2}} {\cos}^3{(\omega_0 t + \phi_0)}$, where the parameters $\phi_0 \in [0, 2 \pi]$, $\omega_0 \in [0.3 \times 2 \pi, 0.6 \times 2 \pi]$ and $a \in [0, 30]$ were sampled from uniform distributions. For each generated signal, we computed the absolute deviation between the ground truth spectrum and those estimated using BFT transform, DFT and tapered DFT. We evaluated the deviations separately for the passband ($\log_{10}g(\xi) > -6$) and the stopband ($\log_{10}g(\xi) < -6$) segments of the spectrum. This division is important since methods that are effective at estimating the main lobes are often poor at suppressing the side lobes and \emph{vice versa}. Since the actual deviations are not informative, we only report the histogram of the ranked performances. Fig.~\ref{figure 1b} shows the results. The BFG transform outperforms both DTF and tapered DTF both in the passband and in the stopband case. As expected, the DTF approach outperforms the tapered DFT approach on the passband range while the opposite is true in the stopband range.

\subsection{Analysis of noise-corrupted signals}
\begin{figure}[!t]
	\centering
    	\includegraphics[width=0.45\textwidth] {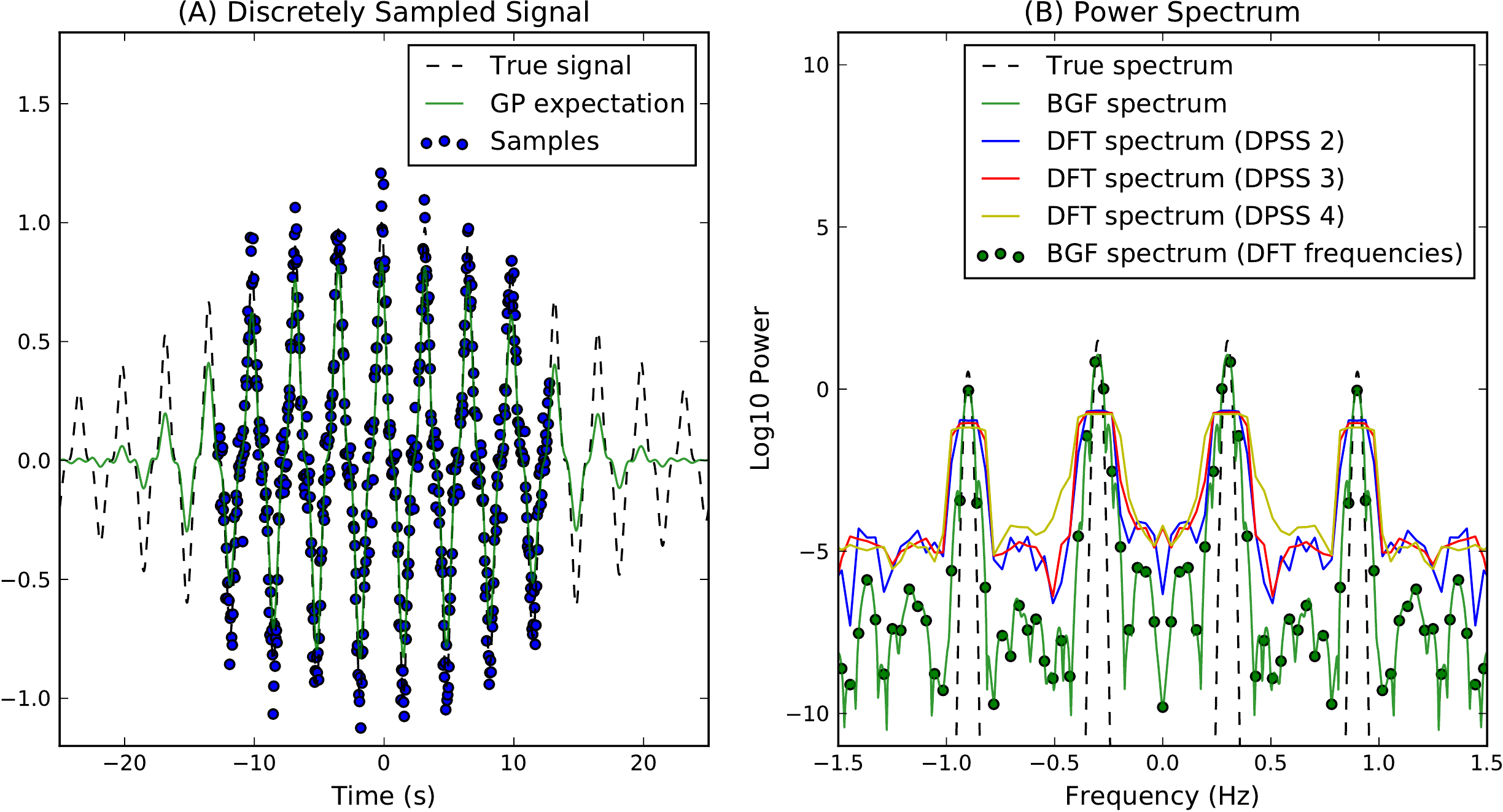}% picture filename
	\caption{Spectral estimation of a synthetic noisy signal. A) Ground truth signal (dashed black line), noise-corrupted sample points (blue dots) and expected value of the GP regression (green line). B) (Log10) Power spectrum of the ground truth signal (dashed line) and spectral estimates obtained using BGF transform (green line), DTF with square taper (blue line) and DTF with two (blue line), three (red line) or four (yellow line) DPSS tapers.}
	\label{figure 2}
\end{figure}

We evaluated the robustness of the method to noise in the time series. As ground truth signal, we used the deterministic signal given in the previous subsection corrupted by Gaussian white noise (sd = $0.1$). We compared the performance of the BGF transform with the performance of a popular multitaper estimator involving discrete prolate spheroidal sequences (DPSS) \citep{percival1993spectral}. We included the DPSS multitaper estimation for this analysis of noisy signals because that method is able to increase the reliability of the noisy estimates by means of spectral smoothing.
Fig.~\ref{figure 2}A shows that the GP expected value acts as a denoiser and remains able to extrapolate the signal beyond the data points. As the noisy data require more regularization, the amplitude of the oscillation is reduced. Fig.~\ref{figure 2}B shows the estimated spectrum. The recovery of the main lobes remains very accurate, apart from a small downward shift due to the amplitude loss. Furthermore, the flat background noise spectrum is more suppressed as compared to the multitaper estimates. 

\begin{figure}[!t]
	\centering
    	\includegraphics[width=0.45\textwidth] {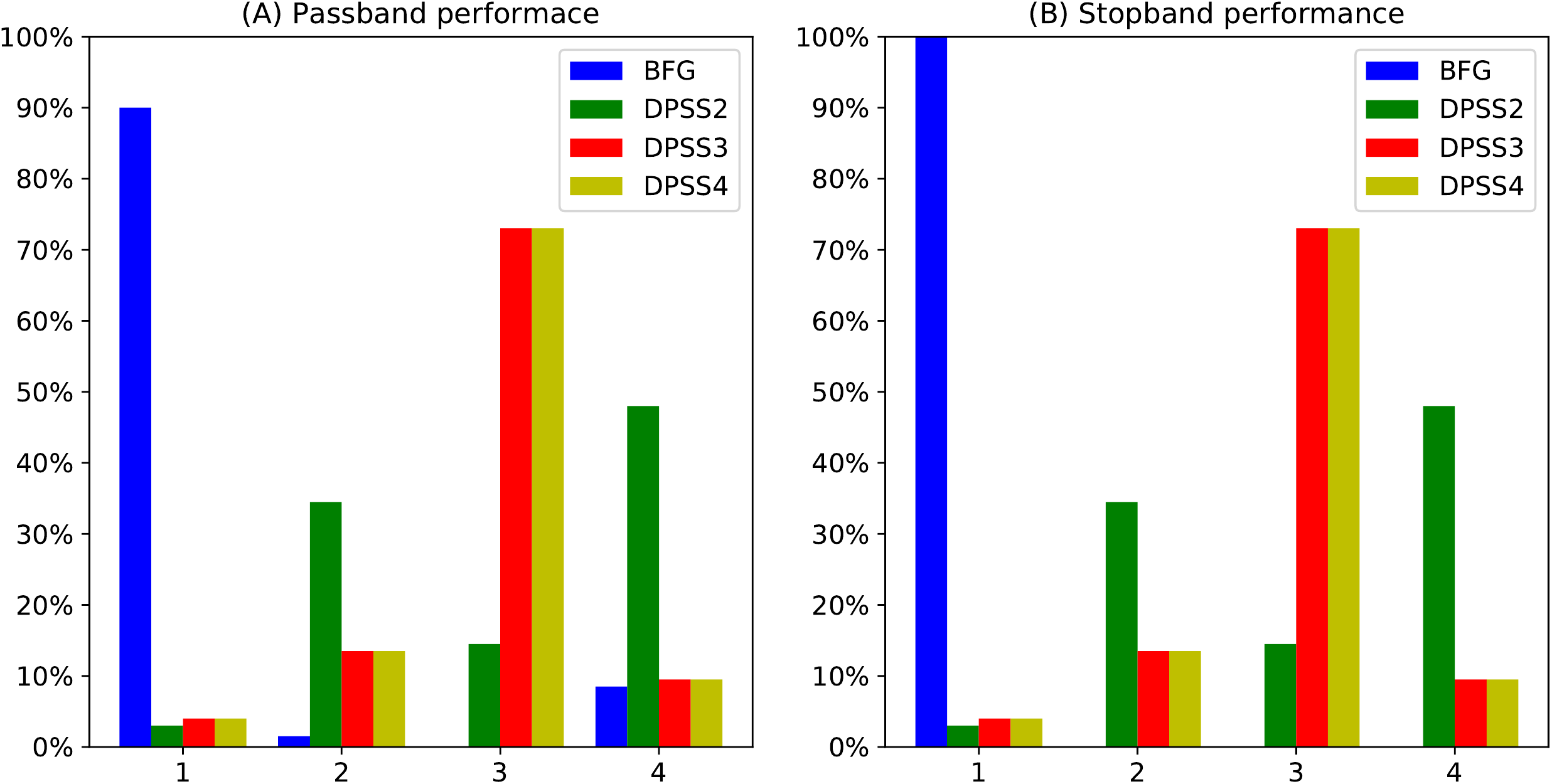}% picture filename
	\caption{Quantitative comparison (noise-corrupted case). Ranked absolute deviations in the A) passband range ($\log_{10}g(\xi) > -6$) and B) stopband range ($\log_{10}g(\xi) < -6$)}
	\label{figure 2b}
\end{figure}

Again, we quantitatively evaluated these observations in a simulation study. The study design was identical to the noise-free case. We corrupted the observation with Gaussian white noise (sd = $0.1$). Fig.~\ref{figure 2b} shows that the BFG transform outperforms all the DFT-DPSS estimates in almost all simulated trials.

\subsection{Fourier analysis of Mauna Loa CO2 level}
\begin{figure}[!t]
	\centering
    	\includegraphics[width=0.45\textwidth] {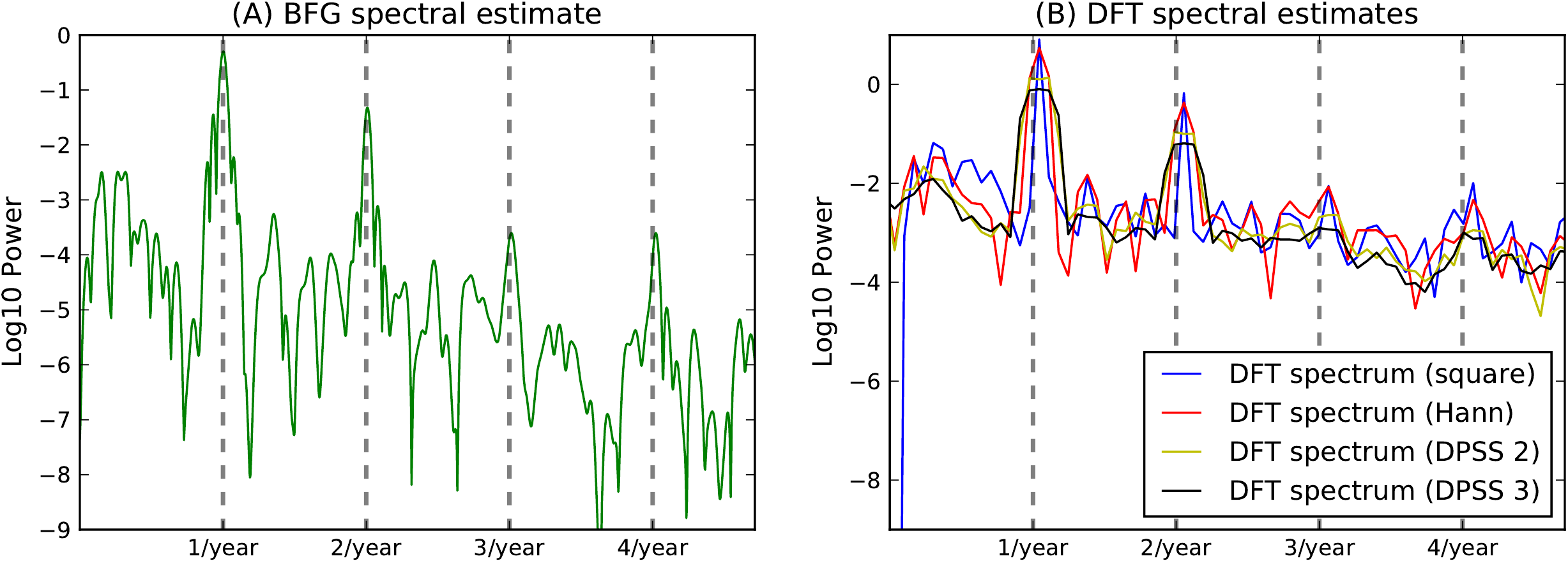}% picture filename
	\caption{Analysis of CO2 level. 1) (Log) Spectral estimate obtained using BGF transform. 2) (Log) Spectral estimate obtained using DTS with square taper (blue), Hann taper (red) and two DPSS tapers (yellow).}
	\label{figure 3}
\end{figure}

As first example using real data, we analyzed the spectrum of the Mauna Loa monthly CO2 concentration. We considered a time period of $15$ years. The time series was de-trended using a second order polynomial regression in order to remove the non-stationary component. We compared the BGF estimate with three DFT estimates well-suited for this noise range: 1) square taper, 2) Hann taper and 3) DPSS tapers (two and three). The results are shown in Fig.~\ref{figure 3}. As we can see, the BFG spectral estimate captures 1) the low frequency broadband component, 2) four sharp peaks corresponding to the one year cycle and 3) the spectral floor. Note that, compared with the DFT-based estimates, the two main peeks are sharper. Furthermore, the peaks corresponding to the 3/year and 4/year frequencies are clearly visible in the BFG estimate but barely discernible in the DFT-based estimates. Tthe energy of the spectral floor is greatly suppressed in the BFG estimate. This effect is probably due to the explicit incorporation of the noise model in the GP analysis. Altogether, the analysis confirms all the features of the BFG transform that we have already established on synthetic signals, namely sharper spectral peeks and higher noise suppression. 

\subsection{Fourier analysis of neural oscillations}
In our final experiment we use the BFG transform to recover the spectrum of neural oscillations. We collected resting state MEG brain activity from an experimental participant that was instructed to fixate on a cross at the center of a black screen. The study was conducted in accordance with the Declaration of Helsinki and approved by the local ethics committee (CMO Regio Arnhem-Nijmegen). Since we are not interested in the spatial aspects of the signal we restricted our attention to the analysis of the MEG sensor with the greatest alpha (10 Hz) power. 
We analyzed the time series using the BGF transform. In this analysis the covariance function of the GP was estimated jointly from all trials by summing the trial specific likelihoods. We compared the resulting spectral estimates with those obtained using DPSS multitaper DFT (with three tapers). Fig.~\ref{figure 3} shows the average and standard deviation of the log-power estimates. The main features of the MEG spectrum are (1) the $1/{f}$ component, a well-known feature of many biological and physical systems \citep{szendro2001pink}, 2) alpha neural oscillations, as visible from the peak at 11Hz and its second harmonic at 22Hz \citep{ward2003synchronous}, and 3) power line noise, sharply peaked at 50 Hz. From the figure we can see that, compared to the DPSS multitaper estimate (panel B), the spectral peaks of the BGF estimate (panel A) are sharper and more clearly visible against the $1/f$ background. 

\begin{figure}[!t]
	\centering
    	\includegraphics[width=0.45\textwidth] {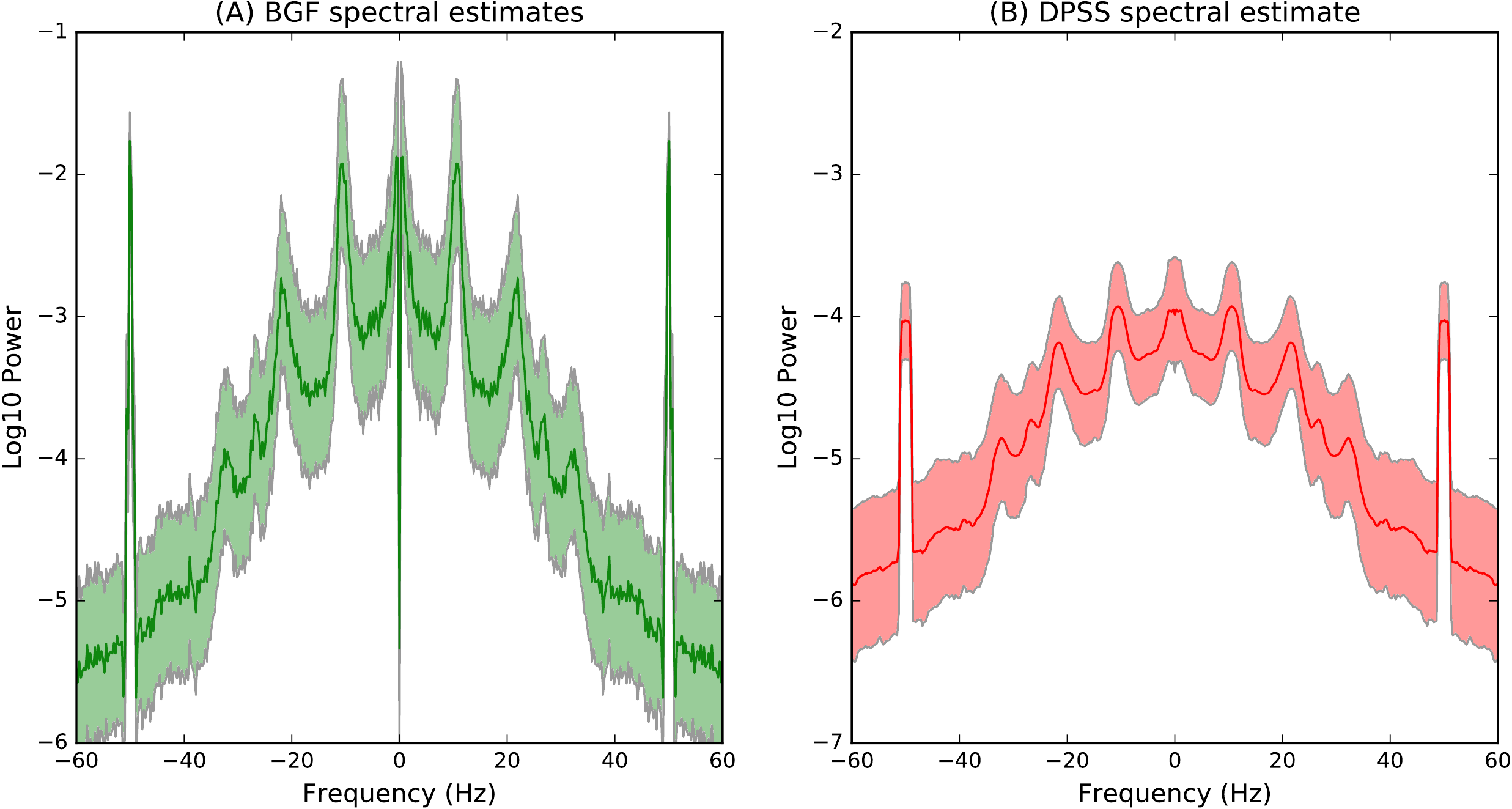}% picture filename
	\caption{Analysis of human MEG signal. 1) (Log) Spectral estimate obtained using BGF transform. 2) (Log) Spectral estimate obtained using DPSS DTS with three tapers.}
	\label{figure 4}
\end{figure}

\section{Conclusions}
In this paper we introduced a new nonparametric Bayesian method for estimating integral transforms of discretely sampled analog signals. While the method can be applied to any linear transform, we focused our exposition on the Fourier transform. We showed that our approach is a probabilistic generalization of the conventional approach based on the famous Nyquist–Shannon sampling theorem. Our Bayesian method depends on the choice of a covariance function. We introduced a new hierarchical Bayesian model which we used in order to estimate the covariance function from the data using a MAP approach. In a series of experiments on simulated and real-world signals, we showed that the resulting BFG transform outperforms the DFT based methods both in terms of mainlobe sharpness and sidelobe suppression.

%\section{Appendix I}
\bibliographystyle{unsrtnat}
\bibliography{BGFbiblio}

\end{document}